\documentclass[letterpaper, 10 pt, conference]{ieeeconf}
\usepackage[hidelinks,colorlinks=true,linkcolor=blue,citecolor=blue]{hyperref}
\usepackage[table,xcdraw]{xcolor}
\usepackage{graphicx}
\usepackage{mathtools}
\usepackage{amsmath}
\usepackage{amssymb} 
\usepackage{amsfonts}
\usepackage{bm}
\usepackage{empheq}
\usepackage{float}
\usepackage{stfloats}
\usepackage{optidef}
\usepackage[absolute,overlay]{textpos}

\IEEEoverridecommandlockouts
\title{\LARGE \bf
Quadrotor Morpho-Transition: \\ Learning vs Model-Based Control Strategies}
\author{Ioannis Mandralis$^{1,2*}$, Richard M. Murray$^{2}$, Morteza Gharib$^{1}$ %
\thanks{Research was conducted at CAST at Caltech}%
\thanks{$^{1}$Aerospace Engineering Department, California Institute of Technology, 1200 E California Blvd, Pasadena, CA, USA {\tt\small mgharib@caltech.edu}}%
\thanks{$^{2}$Computing and Mathematical Sciences Department, Caltech, MA, California Institute of Technology, 1200 E California Blvd, Pasadena, CA, USA {\tt\small mgharib@caltech.edu}
USA. {\tt\small rmurray@caltech.edu}}%
\thanks{$*$ Corresponding author: imandralis@caltech.edu}}

\begin{document}

\newcommand{\blockwidth}{14cm}
\newcommand{\pagecenterX}{3.5cm} 
\begin{textblock*}{\blockwidth}(\pagecenterX, 1cm) 
\centering
\small This paper has been accepted for publication at the 2025 IEEE/RSJ International Conference on\\
Intelligent Robots and Systems (IROS 2025), Hangzhou, China.
\end{textblock*}

\maketitle
\pagestyle{empty}

\begin{abstract}
    Quadrotor Morpho-Transition, or the act of transitioning from air to ground through mid-air transformation, involves complex aerodynamic interactions and a need to operate near actuator saturation, complicating controller design. In recent work, morpho-transition has been studied from a model-based control perspective, but these approaches remain limited due to unmodeled dynamics and the requirement for planning through contacts. Here, we train an end-to-end Reinforcement Learning (RL) controller to learn a morpho-transition policy and demonstrate successful transfer to hardware. We find that the RL control policy achieves agile landing, but only transfers to hardware if motor dynamics and observation delays are taken into account. On the other hand, a baseline MPC controller transfers out-of-the-box without knowledge of the actuator dynamics and delays, at the cost of reduced recovery from disturbances in the event of unknown actuator failures. Our work opens the way for more robust control of agile in-flight quadrotor maneuvers that require mid-air transformation. \href{https://youtu.be/DTHbfR6mCGw}{Video}; \href{https://github.com/mandralis/IsaacLab}{Code}.
\end{abstract}

\section{Introduction}

Ground aerial robotic systems are ideally poised to increase the reliability and scope of autonomous robotic missions. Whilst robots specially adapted to single locomotion types like quadrotors or bipeds may achieve impressive performance in their respective domains, they suffer from fundamental disadvantages. Aerial robots face important limitations due to battery life and payload capacity, whilst ground robots have limited ability to explore the aerial domain. Combining ground and aerial locomotion modes thus promises to deliver increased versatility, better energy efficiency, expanded exploration capabilities, as well as increased fault tolerance.  

\begin{figure}[!t]
    \centering
    \includegraphics[width=0.9\linewidth]{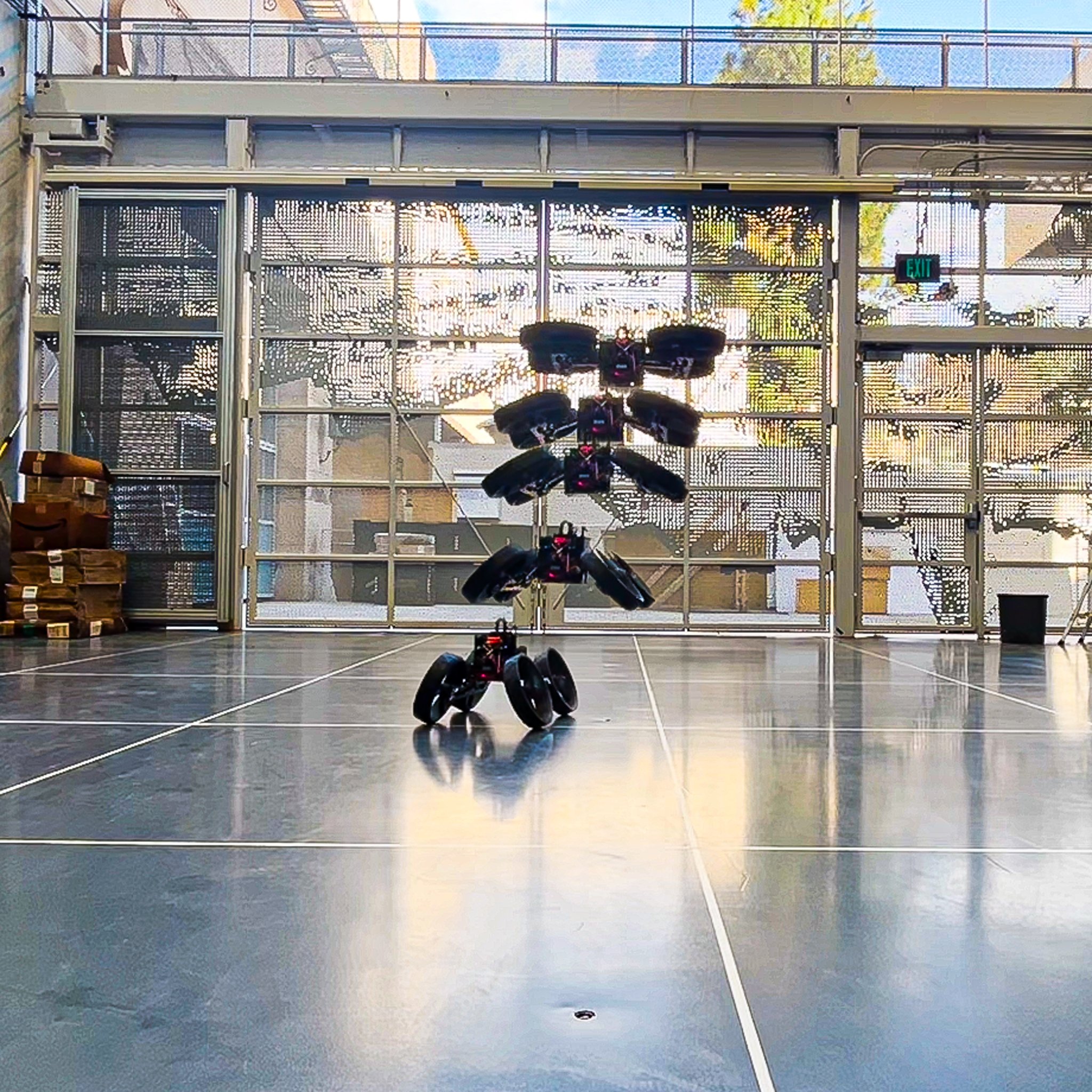}
    \caption{Snapshots of ATMO performing the Morpho-Transition maneuver using an end to end Reinforcement Learning control policy developed in this work. In this maneuver the robot starts in quadrotor flight mode and lands on its wheels by mid-air transformation. }
    \label{fig:1}
\end{figure}

In recent work, a novel ground aerial robot named ATMO (Aerially Transforming Morphobot) capable of \textit{morpho-transition} was introduced \cite{Mandralis2025}. This type of maneuver, depicted in Figure \ref{fig:1} requires smoothly transitioning from flying to driving locomotion by morphing near the ground and landing on dual-purpose wheel-thruster appendages with the largest possible tilt angle, i.e. as close as possible to drive configuration. Contrary to conventional quadrotor landings that involve landing by vertical, non-transforming descent, transforming prior to landing can enable landing and driving in crevices, morphing to enter through narrow tunnels, or enabling smooth ground aerial transition when ground clearance is required due to debris or otherwise rough terrain. 

However, performing such a maneuver can be extremely challenging due to the need to stabilize mid-air while operating near actuator saturation. As ATMO changes its body posture to transform from quadrotor to ground rover mode, it passes through a configuration in which the thrust generated by the spinning rotor blades is not enough to counteract gravity. When attempting to hold position in this configuration, actuators are saturated making disturbance rejection very difficult. Thus, planning through this critical point to land with the largest possible tilt angle, combined with the need to take into account the ground contact upon landing, complicates the design of model based controllers.  

This problem was first studied in \cite{Mandralis2025} by using model predictive control with an adaptive cost function that varies with proximity to the ground and body shape to ensure successful landing on wheels. This method solved the problem of feasibility loss introduced at high tilt angles due to actuator saturation, but required extensive engineering work to achieve the reported results. The method was compared against off-the-shelf cascaded PID quadrotor controllers which were unable to solve the task due to the inherent differences between the dynamics of quadrotor flight and quadrotors in morpho-transition.  

In this work we develop a competing RL method for achieving morpho-transition. The RL agent autonomously learns to stabilize ATMO, reject disturbances, and finds a feasible trajectory to solve the task from a single high level reward objective. Unlike the MPC method, where special care has to be taken when approaching the ground due to the hybrid switching from flight dynamics to ground dynamics, the RL method does not need to explicitly account for ground contact, enabling smooth control throughout the flight-drive operational envelope. The RL method operates at the RPM command level making this an end-to-end approach which bypasses the need for lower level body rate or attitude control loops which require extensive manual tuning, domain knowledge, and engineering effort. 

\begin{figure}[t]
    \centering
    \includegraphics[width=0.75\linewidth]{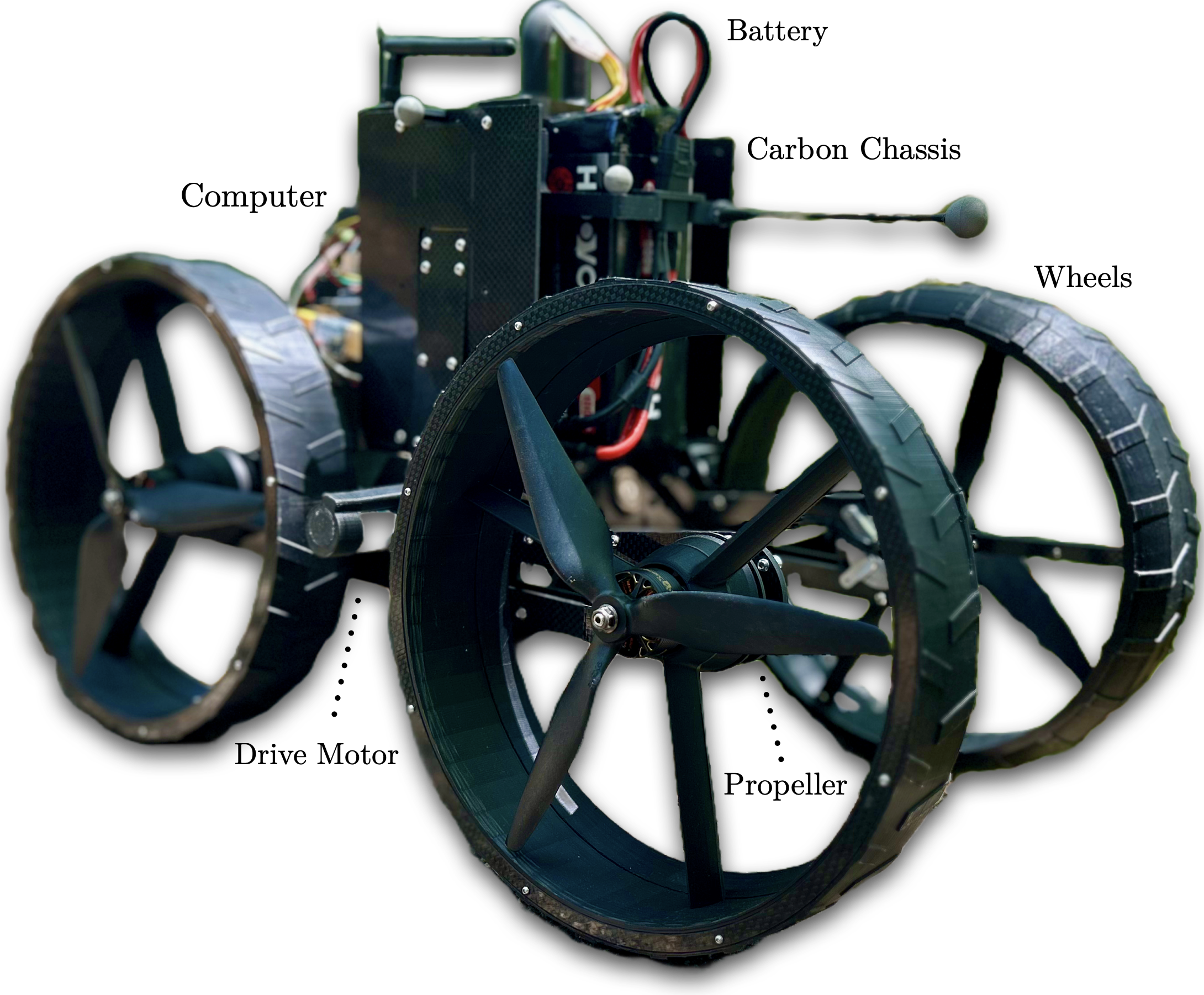}
    \caption{A photo of ATMO, the Aerially Transforming Morphobot, in ground configuration with main components labeled. The robot can drive and fly by virtue of wheel-thruster actuators and a morphing mechanism to change between the two modes.}
    \label{fig:atmo}
\end{figure}

\subsection{Related Work}

Significant work in mid-air transformation control has been done for quadrotors which can move their thrusters in the plane normal to the direction of the thrust. This can be important for quadrotors, for example, which can use this adaptation to fit through narrow gaps, or to optimize flight agility and efficiency. In this space, a feedback linearizing LQR body rate controller is proposed in \cite{Falanga2019} with LQR gains adapted online to take into account the changes of center of gravity and inertia due to shape change. These are then fed into a control allocation matrix, which is a function of the thruster angles, that converts desired torque and collective thrust to thruster commands. This approach is beneficial because it fits directly into the classical cascaded PID loop used for quadrotor control. A competing approach uses an adaptive body rate controller that adapts online to changing actuator configuration \cite{desbiez2017}. 

Fewer works have investigated how rotorcraft with thrusters that tilt away from the vertical axis can be controlled. This type of system poses significant challenges to controllers due to the thrust not being aligned with the body $z$-axis resulting in out-of-plane thrust components that are unaccounted by cascaded PID architectures designed for fixed frame quadrotors. In this domain, there have been developments such as passively morphing out-of-plane quadrotors which use mechanical springs to fold and pass through small gaps \cite{bucki2019,bucki2021}. However, the problem of continuous control of morphing with out-of-plane thrust, rather than hybrid switching, has only been considered using model predictive, or optimization-based control \cite{Mandralis2023, Mandralis2025}. These methods require extensive manual engineering as well as first principles modeling to produce reliable control performance.

An attractive alternative to the model-based techniques is Reinforcement Learning. Research in RL for flight control has mainly focused on learning how to fly quadrotors with the aim of pushing the envelope of performance and agility \cite{Hwangbo2017, Kaufmann2023}. Significant strides have been made in training RL controllers to directly output RPM commands, rather than control commands at a higher level of abstraction, removing the need for manually engineered lower level control loops \cite{Molchanov2019,Kaufmann2022,Eschmann2023,Zhang2024}. Reinforcement learning based landing and perching has been explored for quadrotors \cite{Polvara2017, Kooi2021,Peter2024} but to the best of our knowledge, it has not been used for control of mid-air robotic transformation. 

\subsection{Contribution}
In this work we demonstrate successful transfer to hardware of an end-to-end RL controller and compare our results to an MPC controller for the challenging task of morpho-transition. We show that, using the RL controller, we can achieve controlled landings with tilt angles greater than $65^\circ$, which is challenging to achieve with model-based controllers. We evaluate the performance of the RL method compared to the MPC method in \cite{Mandralis2025} by examining the disturbance rejection characteristics under unknown observation delays, motor dynamics and recovery from partial actuator failure. Our contributions can be summarized as follows:
\begin{enumerate}
    \item We develop a reward function that accurately encodes the problem of morpho-transition by taking advantage of the perfect information available in the simulator.  
    \item We propose a performance benchmark for the morpho-transition maneuver based on impact velocity and final distance to the goal after push disturbances of different magnitude and direction. We thoroughly characterize the controllers by exploiting massively parallel GPU simulation.  
    \item We extend the Isaac Lab simulation framework to include common quadrotor aerodynamics such as the thrust and moment coefficients, as well as motor dynamics and observation delays for the specialized tasks requiring aerial transformation. This code, along with the MPC method, is publicly available \href{https://github.com/mandralis/IsaacLab}{here}. 
\end{enumerate}

\section{Background}

\begin{figure*}[!t]
    \centering
    \includegraphics[width=1\linewidth]{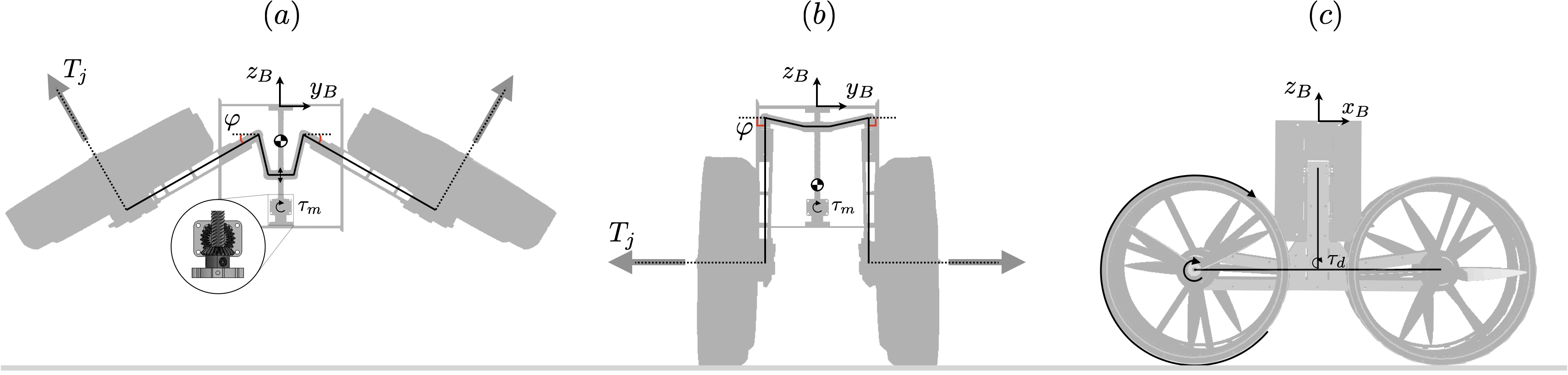}
    \caption{(a) Front-view of ATMO in flight mode with thrust $T_j$ and the body frame depicted. (b) Front-view of ATMO in ground mode. (c) Side-view of ATMO showing the wheel rotation. The tilt angle $\varphi$ is depicted, as well as the motor torques required for morphing, $\tau_m$, and driving, $\tau_d$, respectively.}
    \label{fig:dynamics-modeling}
\end{figure*}

\subsection{Robot Description}

ATMO \cite{Mandralis2025} is an aerially transforming Morphobot capable of flying, driving, and transforming mid-air to land on its wheels. A photo of the robotic platform is shown in Figure \ref{fig:atmo}, along with the main parts and components labeled. The design rationale and system specifications are described in \cite{Mandralis2025}. We provide a brief summary below, for completeness.

ATMO can smoothly transition between quadcopter and ground rover mode by virtue of a closed loop kinematic chain, shown in Figure \ref{fig:mechanism}, which controls the sagittal plane tilt angle, $\varphi$, of four wheel-thruster pairs using a single brushed DC motor. The mechanism offers a high reduction allowing the robot posture to lock in place when un-actuated, and enables transformation with minimal actuation requirements. ATMO's mass of 5.5kg includes a high capacity battery pack for sustained operation, two belt-pulley systems on either side of the robot actuated by DC motors to enable differential drive steering as well as an onboard NVIDIA Jetson Orin Nano computer and CubeOrange PX4 enabled flight controller for sensor fusion and state estimation. All communication is achieved using ROS2 \cite{Macenski2022}.

As ATMO morphs in the sagittal plane to transition from flight to drive modes the thrust produced by the spinning propellers is oriented away from the body-vertical axis resulting in lower overall thrust available to counteract gravity. At the critical tilt angle, $\varphi_c = 60^\circ$,\footnote{Computed as the point where the weight of the robot equals the maximum projection of the thrust in the vertical axis: $\cos\varphi_c = \frac{m g}{T_{\max}}$.} the thruster actuators become fully saturated when trying to counteract gravity. In this configuration, the robot cannot hold its position and reject disturbances simultaneously. Furthermore, when flying in a morphed configuration, unequal values of thrust on either side of the H-frame rotor configuration may result in thrust components along the body $y_B$-axis, shown in Figure \ref{fig:dynamics-modeling}. This makes ATMO's dynamics fundamentally different to a fixed-frame quadrotor. 

\subsection{Morpho-Transition Quadrotor Dynamics}

Following \cite{Mandralis2025}, we modeled ATMO as 7 inertial components: the robot base, the left arm and right arm, and the four spinning propellers, as shown in Figure \ref{fig:dynamics-modeling}. The wheel joints are not considered for aerial maneuvers. 

The robot generalized coordinates and velocities are defined as: $\bm q = (\bm p,\bm \xi, \varphi)$ and $\bm w =(\bm v, \bm \omega)$ where $\bm p$ is the position of the robot center of mass, $\bm \xi$ is the quaternion of the robot base frame relative to the inertial world frame, $\varphi$ is the body tilt angle, $\bm v$ is the velocity of the center of mass, and $\bm \omega$ is the angular velocity of the robot relative to the world frame expressed in the body frame. 

The control inputs are defined as $\bm u = (\bm u_{\textrm{aero}}, u_{\textrm{body}})$, where $\bm  u_{\textrm{aero}} \in [0,1]^4$ denote the normalized propeller RPMs and $u_{\textrm{body}}\in[-1,1]$ is the commanded tilting velocity. The thrust force and drag torque caused by the rotating propellers are assumed to be linear functions of the control inputs and act parallel to the rotor axes which are defined in the positive body $z$ axis: $\bm T_j = c_T  u_{\textrm{aero}}^j \bm {\hat z}_j$ and $\bm M_j = c_M c_T  u_{\textrm{aero}}^j \bm {\hat z}_j$. The thrust and moment coefficients $c_T, c_M$ are identified experimentally through thrust stand system identification experiments. $\bm {\hat z}_j$ denotes the rotation axis of propeller $j$.

The body shape change occurs by spinning a central shaft with a DC motor encoder combination which results in linear motion of joint $A$, as shown in Figure \ref{fig:mechanism}. This in turn is translated to a rotation of point $E$, which corresponds to the intersection of the propeller axis with the link $CDE$. Thus a single actuator can transform the robot. This effect is modeled as a pure integrator for the tilt angle $\varphi$, reflecting the physical properties of the non-compliant morphing mechanism. The motor dynamics of the thrusters are modeled as first order linear systems with input $\bm u_{\textrm{aero}}$ and output the normalized propeller RPMs ${\bm \Omega}$. 

Overall, ATMO's dynamic equations of motion can be written in the following form, 
\begin{empheq}[left = \empheqlbrace]{align}
    &\bm M(\bm q)\bm{\dot w} + \bm b(\bm q,\bm w) + \bm g(\bm q) = \bm S(\bm q) \bm \Omega \\
    &\dot {\bm \Omega} = T_m^{-1} (\bm u_{\textrm{aero}} - {\bm \Omega}) \\
    &\dot \varphi =  \dot \varphi_{\max} u_{\textrm{body}},    
\end{empheq}
where $\bm M(\bm q)$ denotes the mass matrix, $\bm b(\bm q, \bm w)$ encodes the coriolis terms, and $\bm g (\bm q)$ encodes the gravitational dynamics. The actuation matrix $\bm S(\bm q)$ maps the generalized control input into the generalized acceleration space. The thruster motor constant is $T_m$ and the maximum speed of the closed loop kinematic linkage is $\dot \varphi_{\max}$. The robot equations were derived using the open-source software package proNEu \cite{Hutter2011}. In state-space, the dynamics take the following form:
\begin{align}
    \bm{\dot{x}} &= \bm f(\bm x,\bm u), \\
    \bm x &=(\bm p,\bm \xi, \bm v, \bm \omega, \varphi, \bm \Omega)\in \mathbb{R}^{18}, \\
    \bm u &= (\bm  u_{\textrm{aero}},  u_{\textrm{body}})\in [0,1]^5.
\end{align}
Second order aerodynamic effects like proximity effects, propeller flapping, or vehicle rotor dynamic couplings \cite{Bristeau2009} are ignored.

\subsection{Model Predictive Controller}
The model predictive controller for the morpho-transition maneuver is detailed in \cite{Mandralis2025} and briefly summarized here. The controller is a trajectory tracking MPC that receives a desired reference $x_\textrm{ref},u_\textrm{ref}$ and solves the following optimal control problem iteratively in a receding horizon control loop:
\begin{mini*}|1|[2]
        {\bm x,\bm u}
        {\int_0^{t_f} L(\bm x, \bm u) d t}
        {\label{eq:cost}}{}
        \addConstraint{}{\bm {\dot{x}} =  \bm f(\bm x,\bm u),}{\forall t\in[0,t_f]}
        \addConstraint{\bm  u_{\textrm{aero}} \in [0,1]^4. \quad }{}{}
    \end{mini*}
The dynamics are discretized and enforced using multiple shooting \cite{Bock1984} with $N$ collocation points and a look-ahead time $t_f$.  A nonlinear program is solved using sequential quadratic programming in a real-time iteration scheme onboard the NVIDIA Jetson Orin Nano using the software packages CasADi and ACADOS \cite{Verschueren2019acadosaMO,Andersson2018}. The controller runs at 150Hz. A look-ahead of $t_f = 2.0$ seconds and $N=20$ collocation points are used. The cost function is given by,
\begin{align}
    L(\bm x, \bm u) = \alpha(\bm x) \underbrace{L_1(\bm x, \bm u)}_{\textrm{flying}} + (1 - \alpha(\bm x)) \underbrace{L_2(\bm x, \bm u)}_{\textrm{transitioning}},
\end{align}
and is a convex combination of quadratic costs $L_1,L_2$ specialized for flight and transition \cite{Mandralis2025}. The cost is varied online according to the height and body shape through the scalar function $\alpha(\bm x)$. The MPC controller assumes that the RPM commands $\bm \Omega$ are equal to the control commands i.e. $\bm u_{\textrm{aero}} = \bm \Omega$, thus operating in the reduced state space $\bm x = (\bm p,\bm \xi, \bm v, \bm \omega)$. The tilt angle $\varphi$ is supplied from an external reference generator and tracked by a low level controller. For details on the implementation of the MPC algorithm, the reader is referred to \cite{Mandralis2025}.

\section{Reinforcement Learning Approach}

\subsection{Simulation}

To train a control policy for morpho-transition maneuvers, we implement morphing quadrotor dynamics as an extension of the state-of-the-art Isaac Lab simulation framework \cite{makoviychuk2021,mittal2023orbit}. The following section gives a brief overview of the dynamics implemented in the simulator. 
ATMO is simulated as a rigid body using a URDF (Unified Robot Description Format) representation with kinematic and inertial parameters generated from a CAD model. The simplified dynamics models for the closed loop kinematic linkage, aerodynamics models, and the motor dynamics are implemented in discrete time in the pre-physics step of the simulation, which performs the following operations:
\begin{empheq}[left = \empheqlbrace]{align}
    \bm \Omega [k+1] &= \beta \bm u_{\textrm{aero}}[k]  + (1 - \beta)\bm \Omega[k]\\
    \bm T_j[k+1] &= c_T  \bm \Omega_j[k+1] \bm {\hat z}_j \\
    \bm M_j[k+1] &= c_M^j c_T \bm \Omega_j [k+1] {\bm {\hat z}}_j\\
    \varphi_d[k+1] &= \varphi [k] + {\dot \varphi}_{\max} u_{\textrm{body}}[k] T_s.
\end{empheq}
The motor dynamics are discretized exactly, under the assumption of zero-order hold between simulation time-steps, leading to a first-order filter with constant $\beta = 1 - e^{-T_s / T_m}$. A nominal motor time constant of $T_m = 0.15$ is used, and the simulator time step is $T_s = 0.02$ seconds, leading to a filter constant $\beta = 0.12$. The filtered normalized motor RPMs are then used to determine the thrust and moment acting on the four propeller axes, $\bm T_j, \bm M_j$. 

\begin{figure}[!t]
    \centering
    \includegraphics[width=1\linewidth]{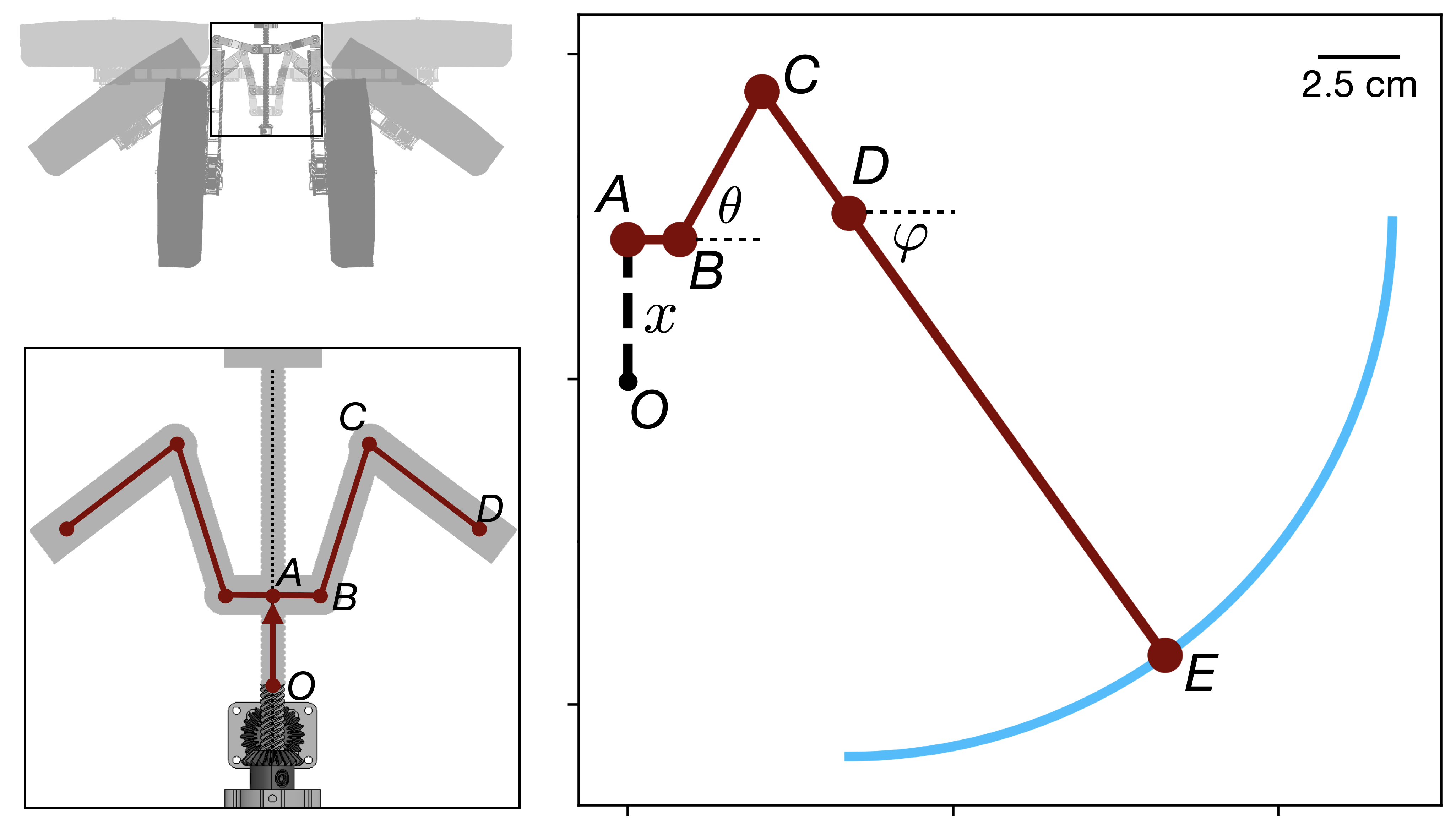}
    \caption{A depiction of the closed loop kinematic linkage enabling symmetric morphing and transformation from air to ground mode.}
    \label{fig:mechanism}
\end{figure}

The desired tilt angle $\varphi_d$ is given to the simulator and tracked by two PD position controllers (one for each arm joint), with stiffness $k_p = 1\times 10^{15}$ and damping $1\times 10^5$. The actual tilt mechanism on the hardware, depicted in Figure \ref{fig:mechanism}, is a closed loop kinematic chain that is difficult to simulate \cite{rudin2022,el-agroudi2024}. Instead, we simplify the actuation system to a ``stiff" mechanism driven by a single velocity input. The large gains chosen for the PD controller make the actuator a pure velocity controller that ensures the mechanism is stiff - approximating its real world characteristics. 

\begin{figure*}[t]
    \centering
    \includegraphics[width=0.70\linewidth]{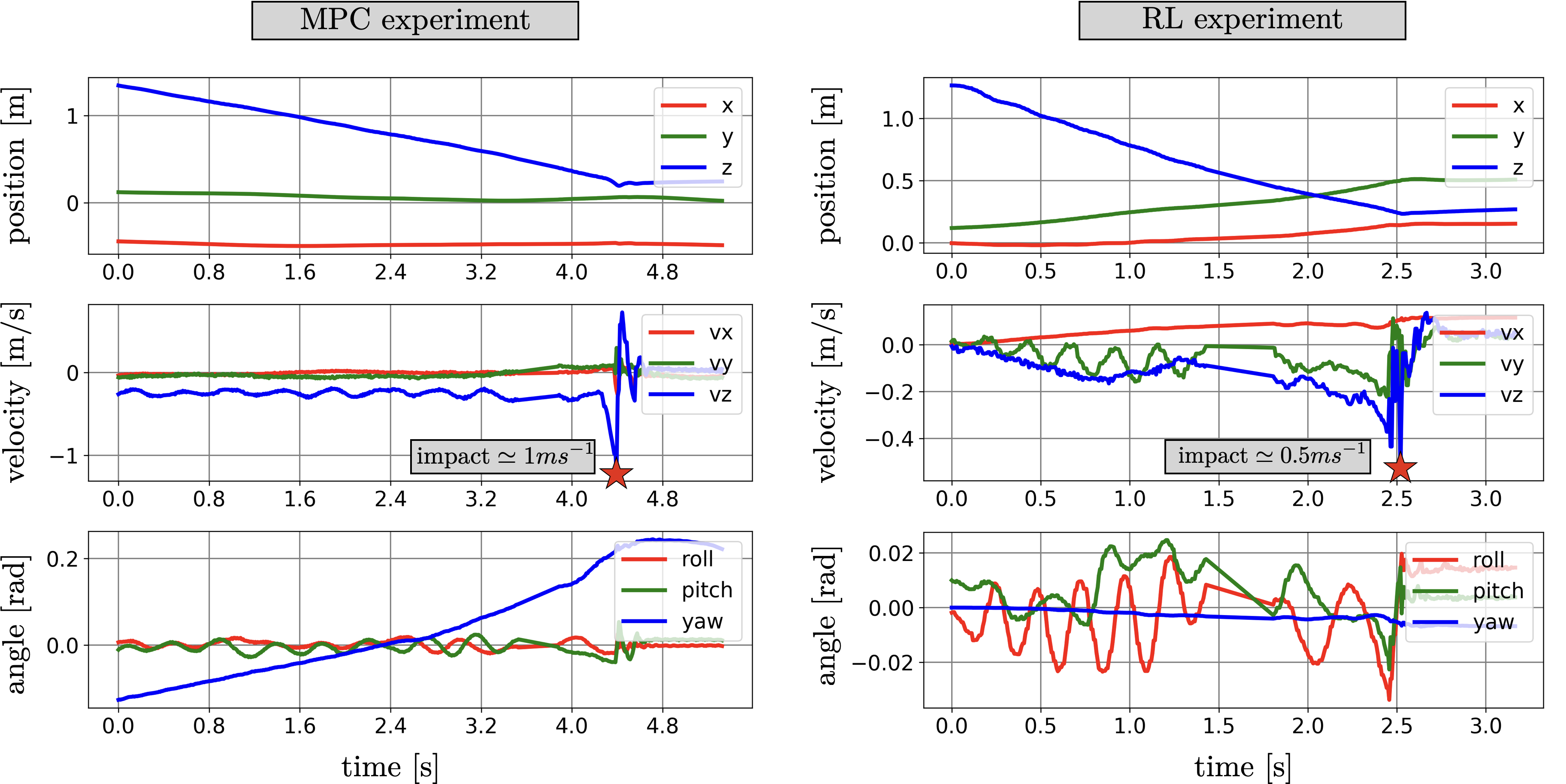}
    \caption{Morpho-Transition maneuver performed on the hardware using the two controllers. The performance of the MPC controller is shown on the left, and the RL controller on the right. The key states during the maneuver are reported.}
    \label{fig:experimental-results}
\end{figure*}

\subsection{Task Definition \& Domain Randomization}
At the beginning of each episode ATMO spawns uniformly at a random $(x_0,y_0)$ position in a square of 2 meters side length around the origin, and an initial height, $z_0 \sim \mathcal{U}(1,2)$ meters from the ground. The goal state is set as the origin. The initial orientation relative to the inertial frame is randomized by sampling a roll and pitch from $\mathcal{U}(-\frac{\pi}{6},\frac{\pi}{6})$ and an initial yaw angle $\psi_0$ from $\mathcal{U}(0,2\pi)$. Initial velocities, $\bm v_0$, and angular velocities, $\bm \omega_0$ are sampled uniformly from the distribution $\mathcal{U}(-0.1, 0.1)$. The initial tilt angle is sampled randomly as $\varphi_0 \sim \mathcal{U}(0,\pi/6)$. Each episode ends when the elapsed simulation time exceeds five seconds, when the robot speed exceeds a maximum threshold of $2.0 m s^{-1}$, or the $xy$ position of the robot deviates from the goal position by more than $1.5$ meters.

We make the policy robust to disturbances by pushing the robot in random directions once per episode. At the beginning of each episode a push time, $t^\ast$, and push duration, $T^\ast$, are sampled from uniform distributions $t^\ast \sim \mathcal{U}(0,0.8 t_f), T^\ast \sim \mathcal{U}(0,0.2)$, where $t_f$ is the episode length i.e. 5 seconds. The force and moment direction and intensity are sampled at the beginning of each episode and applied in the pre-physics step of the simulator for the given duration at the push time. The force and moment scales are set to $f^\ast = 0.20 c_T$ and $\tau^\ast = 0.20 c_T c_M$. The maximum impulse and angular impulse imparted on the robot during training is thus: $J_{\textrm{disturbance}}^{\max} = 0.04c_T$ and $\Delta L_{\textrm{disturbance}}^{\max} = 0.04 c_T c_M$.

The motor time constant is randomized slightly around the nominal value $T_m = 0.15$ seconds at the beginning of each episode $T_m \sim \mathcal{U}(0.10,0.20)$. The thrust and moment constants are assumed to have a $\pm 20\%$ uncertainty and are sampled as $c_T \sim \mathcal{U}(0.8 c_T,1.2 c_T)$ and $c_M \sim \mathcal{U}(0.8 c_M,1.2 c_M)$. Finally, the uncertainty of the tilt actuator dynamics is incorporated by randomizing the maximum tilt velocity parameter $\dot \varphi_{\max} \sim \mathcal{U}(0.8\frac{\pi}{8},1.2\frac{\pi}{8})$. We found that training without randomizing these parameters results in neither sim-to-sim nor sim-to-real transfer. 

\subsection{Action and Observation Spaces}
The actions are the outputs of the trained RL policy, $\boldsymbol \pi(\boldsymbol s)$. For the policy network we use three hidden layers of 128 units each with Exponential Linear Unit (ELU) activation functions between the layers. The network outputs are passed through a sigmoid layer to ensure the control actions are $\boldsymbol u = \boldsymbol \pi(\boldsymbol s) \in [0,1]^5$. The outputs of the policy network are interpreted as the 4 thruster control inputs $\bm u_{\textrm{aero}}$ and the tilt angle velocity input $u_{\textrm{body}}$. 

The inputs to the policy network are the policy observations,
\begin{equation}
    \boldsymbol s_{\pi} = (\boldsymbol p,\boldsymbol R,\boldsymbol v,\boldsymbol \omega,\varphi, \bm u^{-}) \in \mathbb{R}^{19 + 5n},
\end{equation}
where $\bm u^{-}$ is an observation history of $n=10$ previous time steps. For the rotation representation we found that it was necessary to use the full rotation matrix $\boldsymbol R \in SO(3)$. Using a Quaternion representation of rotation did not result in sim-to-real transfer since quaternions double cover the space of rotations, meaning that the policy network must learn that $\bm \xi$ and $-\bm \xi$ represent the same rotation. 

We use an asymmetric actor critic  \cite{pinto2017} scheme and give 14 additional privileged observations to the critic to improve value estimation. The critic observation space thus becomes,
\begin{equation}
    \boldsymbol s_Q = (\boldsymbol s_{\pi}, \boldsymbol f, \boldsymbol \tau, t^\ast, T^\ast, t, J_c, \boldsymbol \Omega^{-})  \in \mathbb{R}^{33+5n},
\end{equation}
where $\boldsymbol f, \boldsymbol \tau$ are the disturbance force and moment, $t^\ast, T^\ast, t$ are the push time, push duration, and current time, and $J_c$ is the impulse acting on the robot due to ground contact forces. This is approximated as, $J_c = \int_t^{t+T_s} \boldsymbol f_c d t \simeq \left[\boldsymbol f_c(k+1) - \boldsymbol f_c(t)\right] T_s$, where $\bm f_c$ is the contact force between the ground and the robot, obtained from the Isaac Lab simulation environment. We gave the critic network access to the observed RPM values $\bm \Omega^{-}$ which was shown in \cite{Eschmann2023} to stabilize training. A small amount of uniform noise, with scales available in the Appendix, is added to the policy observations. No noise is added to the action history. 

\subsection{Delays \& Motor Dynamics}
Finally, we found that for successful sim-to-real transfer, it was essential to add an observation delay of one simulation time step, or $20ms$, to the observations fed into the policy network during train time. This approximates the delays present due to inter-computer communication on the hardware. Without taking into account observation delays as well as motor dynamics the RL control policy could not be transferred to hardware.

\begin{figure*}[t]
    \centering
    \includegraphics[width=0.75\linewidth]{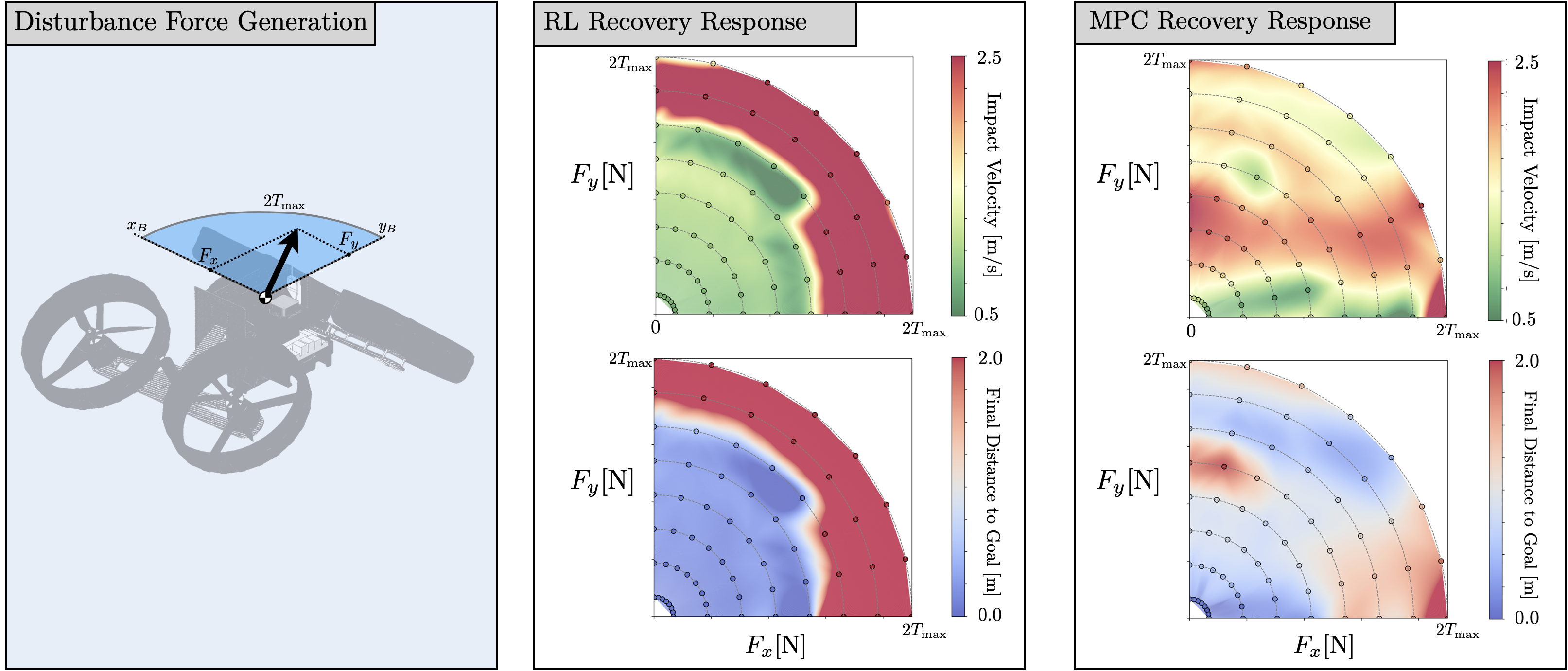}
    \caption{Comparison between RL and MPC methods. The impact velocity and final distance to goal was recorded for 64 different push forces for each controller. The direction of the push force was varied in the $xy$ plane and the magnitude of the push force was varied between $0.60 c_T$ and $8.0 c_T$. The data points are shown in a scatter plot with an interpolated heat map overlayed.}
    \label{fig:disturbance}
\end{figure*}

\subsection{Reward Function}
The reward function is designed to incentivize landing on the wheels in the vicinity of the goal state. The main term that produces this behavior is: $r_c^{w} = (c_{w} \wedge a)$. Here, $c_w$ is a Boolean variable active when any of the wheels make contact with ground and $a$ is a Boolean variable denoting that the robot is in the acceptance state, defined as the state where the $xy$ position of the robot is within 40 cm from the goal position.

To ensure that the robot has sufficiently rich reward information we added various reward shaping terms. First, we penalized the linear and angular velocities of the robot base frame, the action rate, and deviations from flat orientation. We rewarded the distance to the $xy$ coordinates of the goal position as well as descending at a constant rate of $0.5 ms^{-1}$. We also rewarded high tilt angles as the robot approaches the ground. Finally, we took advantage of the ability to accurately measure contact times and forces in the simulator to penalize large impulse forces with the ground as well as to penalize undesirable thrust actions that occur while the robot is in contact with the ground. We impose a penalty of $-1$ unit for early termination. The full reward function expression is:
\begin{align}
   r(\bm s, \bm u) = &a_0||\bm v||^2 +a_1||\bm \omega||^2 +a_2|1-q_a|\nonumber \\
   &  +a_3||\bm u - \bm u^{-}||^2 + a_4 c_0||\bm u_{\textrm{aero}}||^2 \nonumber \\
   & +a_5 J_c^2 +a_6(c_w\wedge a)\nonumber \\
   & +a_7 \phi(d) + a_8\phi(z^2 + e_{\varphi}^2) \nonumber \\
   & +a_9\phi(e_{v_z}^2).
\end{align}
Here, $\phi(\cdot)$ is the function $\phi(x) = e^{-4x}$. $q_a$ denotes the axis of the quaternion, $c_0$ is a boolean variable that is equal to 1 when the first contact with the ground has occurred, $d$ is the distance to the $xy$ position of the goal state, and $e_{\varphi} = \varphi - \frac{\pi}{2}$. The coefficients $a_1\dots a_9$ are given in the appendix. 

\subsection{Training}
We use the Proximal Policy Optimization algorithm (PPO) \cite{schulman2017} and exploit massively parallel training on GPU \cite{rudin2022}. We use the RL-games implementation \cite{makoviychuk2021} with an initial learning rate of $\lambda=1\times 10^{-5}$. Our model was trained for 1000 policy updates after which the reward had converged to a stable value. This took about 20 minutes on a GPU-enabled desktop computer.\footnote{NVIDIA GeForce RTX 4070 with 16GB memory}

\begin{figure}[t]
    \centering
    \includegraphics[width=1\linewidth]{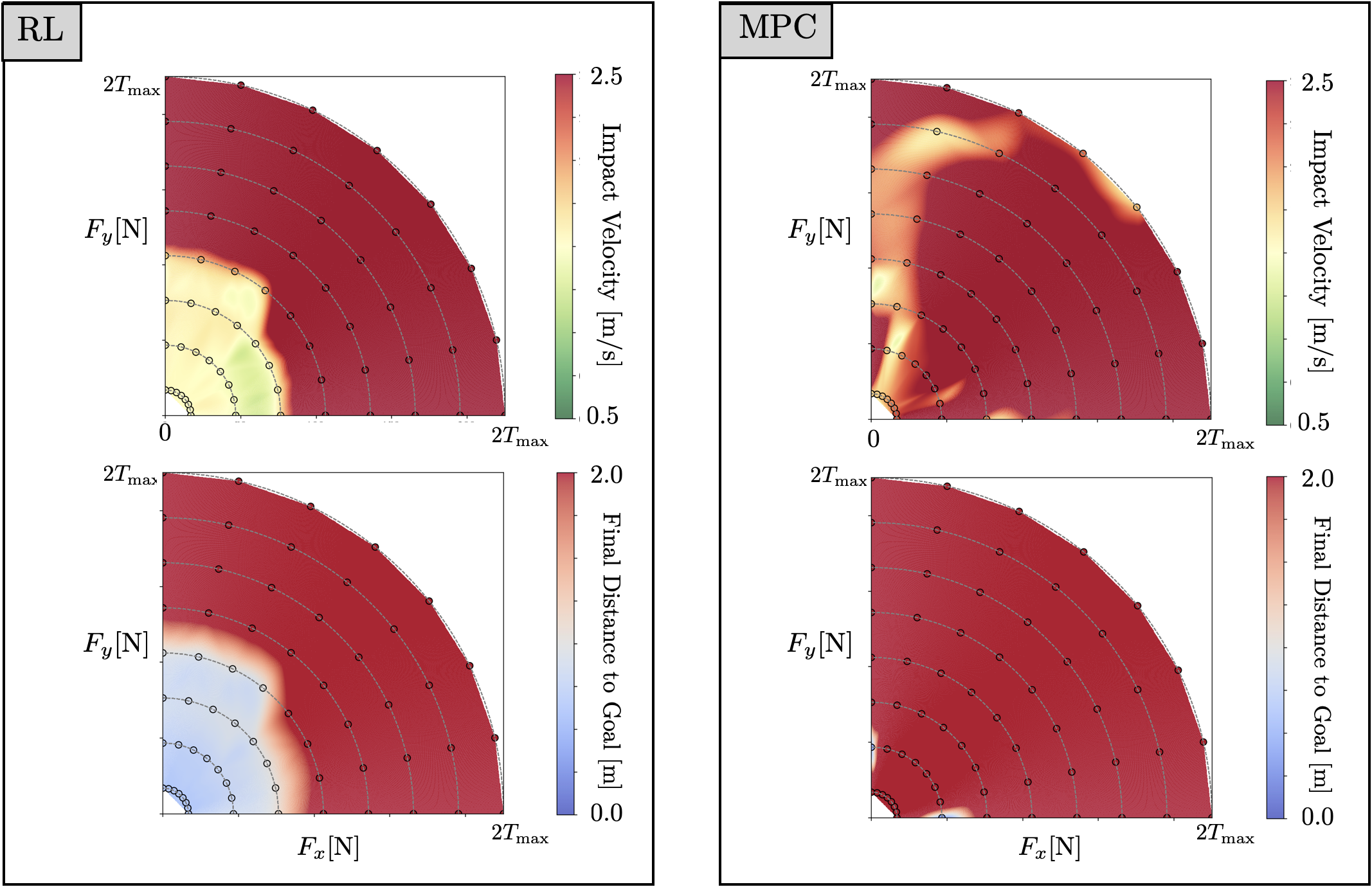}
    \caption{Recovery characteristics under partial actuator failure. The thrust and moment coefficients of the thrusters are multiplied by  $[0.8,0.9,0.85,1.1]$ with order front right, back left, front left, back right.}
    \label{fig:actuator-failure}
\end{figure}

\section{Hardware Experiments}

We performed different Morpho-Transition maneuvers in the CAST flight arena at Caltech, using the MPC and the RL controllers. Optitrack motion capture information was used to supply position and orientation estimates to an Extended Kalman Filter running onboard ATMO removing the need for GPS or vision based localization. ATMO was initialized on the ground and flown up to 1.25 meters using the PX4 quadrotor position controller \cite{Meier2015}. At that point, the RL or MPC controllers were engaged and the morpho-transition landing maneuver was performed. The results of two representative experiments are shown in Figure \ref{fig:experimental-results}, and the videos are available as supplementary materials.

Both controllers were successfully able to complete the task. The final tilt angle achieved by the MPC controller was $\varphi=60^\circ$ and the RL controller achieved $\varphi=65^\circ$. The full maneuver can be seen in the Supplementary Video accompanying this paper. The RL controller achieves better impact velocity $0.5m s^{-1}$ as opposed to the $1.0m s^{-1}$ impact achieved by the MPC controller. It also manages to do this while operating past the actuator saturation tilt angle. 

On the other hand, the RL controller exhibits larger oscillations in the roll angle evidencing slight instability during the descent. We believe this is due to inaccurate estimation of the system latency as well as the motor dynamics time constant. The effect of delays can be mitigated by inferencing the RL control policy directly on the flight controller micro-controller hardware rather than on the companion computer which adds latency due to the need to stream commands through the ROS2 network. Overall, the MPC method exhibits more stable angular dynamics but shows significant drift in yaw. 

\section{Recovery from Disturbances}

To further characterize the differences between the two methods, we exploited the parallel simulation environment of Isaac Lab to examine the effect of the direction and magnitude of push disturbance on both controllers. 

The MPC method is tasked with following a downward descending trajectory at 0.5 m/s and the RL agent performs actions according to the policy learned during the training process. The simulation was set up with $T_m=0.15$ and $20ms$ observation delay. We systematically applied a push force, in the $xy$ plane of center-of-mass frame, at different angles and magnitudes in the first quadrant of the body $xy$ plane, as shown in the first panel of Figure \ref{fig:disturbance}. The angle was varied in increments of $\frac{\pi}{16}$ in the first quadrant of the body $xy$ plane. ATMO's symmetry does not require us to test push forces in other quadrants. The magnitude of the force was varied in increments of $0.925 c_T$ from $0.6 c_T$ to $8.0 c_T$. Thus the maximum disturbance force that the robot recovery was tested for was of magnitude 2 times greater than the overall thrust capability of the robot ($T_{\max} = 4 c_T$). 

To characterize the ability of the controllers to recover from push disturbances we employ two metrics: 1) the impact velocity, and 2) the final distance from the goal. For each applied disturbance force the control policies were rolled out in simulation for 5 seconds and the impact velocity and final distance to the goal were recorded. Lower impact velocity and lower final distance to goal indicate that the controller was able to successfully recover from the push disturbance. Our results are shown in Figure \ref{fig:disturbance}. In the case of the RL method, the control policy performed very well up to between $5-6 c_T$ disturbance force. In this region, there is a low impact velocity close to the commanded $0.5 ms^{-1}$. The final distance to goal also stays close to zero. In other words, the RL agent is able to fully recover from push disturbances which are far beyond what it has been trained on. Once the threshold of $5-6 c_T$ is surpassed however, the RL agent fails to land with an acceptable impact velocity or final distance to goal. 

The MPC controller, on the other hand, exhibits more uniform performance. It is able to recover reasonably well from larger disturbances but it does not exhibit a uniform region of excellent disturbance recovery performance. Indeed, although the MPC controller often arrives at the goal position, it also often impacts the ground with unacceptable impact velocity.

\section{Recovery from Partial Actuator Failure}

Finally, we tested the performance of the two controllers under partial thruster failure. This can happen in experimental scenarios due to motor wear or low energy state. To simulate this scenario, we multiplied the nominal value of the thrust and moment coefficients by $[0.8,0.9,0.85,1.1]$. This is a significant perturbation that the RL has not been trained on\footnote{the RL was trained only on uniform changes in the thrust and moment coefficients.}, and the MPC has no knowledge of. The results are shown in Figure \ref{fig:actuator-failure}. The RL controller is able to recover from considerable disturbance forces, even under the partial actuator loss. Naturally, the recovery region is much smaller than in the case where the thrusters are operating at nominal efficiency, Figure \ref{fig:disturbance}, but the RL agent is still able to recover from considerable applied impulses. In contrast, the MPC controller cannot recover at all, resulting in failure for almost all push tests considered. Note that it is possible to extend the MPC algorithm to take into account actuator failure as done in \cite{Nan2022}, which is expected to significantly improve performance. However, this would require explicit estimation of the actuator failure to trigger online changes in the optimal control problem being solved by the MPC.

\section{Conclusions \& Future Work}

We have demonstrated successful transfer of an end-to-end deep RL controller to hardware for the task of quadrotor morpho-transition. To achieve this we developed a training approach that can generalize for the control of aerial robots in need of mid-air transformation control. The RL approach was compared to an MPC method. We found that end-to-end RL that has been trained on a well informed distribution of disturbance dynamics can reject small disturbances more reliably than the equivalent end-to-end MPC method. The MPC method performs worse at small disturbances but is able to recover from large disturbances where the RL controller fails. Moreover, the RL method is able to recover from partial actuator failure without explicit knowledge of the failure. The MPC method can in theory be extended to achieve this but would require explicit estimation of the actuator failure to trigger online changes in the optimization. This highlights the potential of RL to generate control policies that can generalize from partial sensor information to produce sophisticated control behaviors. In future work, we will seek to characterize the disturbance rejection capabilities of the two controllers in experiment, and investigate how an MPC approach that takes motor dynamics and observation delays into account compares to the end-to-end RL control policy. 


\addtolength{\textheight}{0cm}   
\section*{ACKNOWLEDGMENT}

Gabriel Margaria carried out preliminary work on RL-based morpho-transition which greatly contributed to this project. This paper benefited from numerous insightful discussions with Adrian Ghansah and Aaron Ames, who encouraged the comparison of the RL controller to the MPC method. We thank Joshua Gurovich for help with experiments. We would like to acknowledge funding from the Center for Autonomous Systems and Technology (CAST) at Caltech. IM is an Onassis Scholar.

\bibliographystyle{IEEEtran}
\bibliography{references.bib}

\section{Appendix}
\subsection{Reward function coefficients}
\begin{table}[ht]
    \centering
    \begin{tabular}{|>{\columncolor[gray]{0.9}}c|c|c|c|}
        \hline
        \textbf{Coefficient} & \textbf{Value} & \textbf{Description} \\ \hline
        $a_0$ & $-0.10 \, T_s$ & Velocity Penalty \\ \hline
        $a_1$ & $-0.30 \, T_s$ & Angular Velocity Penalty \\ \hline
        $a_2$ & $-0.10 \, T_s$ & Orientation penalty\\ \hline
        $a_3$ & $-0.07 \, T_s$ & Control Rate Penalty \\ \hline
        $a_4$ & $-0.13 \, T_s$ & Ground Thrust Penalty \\ \hline
        $a_5$ & $-1.0$ & Contact Impulse Penalty \\ \hline
        $a_6$ & $0.40$ & Contact in Acceptance Reward \\ \hline
        $a_7$ & $0.30 \, T_s$ & Distance to Goal Reward \\ \hline
        $a_8$ & $0.40 \, T_s$ & Morphing reward \\ \hline
        $a_9$ & $0.30 \, T_s$ & Descending reward \\ \hline
    \end{tabular}
    \label{tab:reward_definitions}
\end{table}

\subsection{Noise scales}
We added uniform noise of magnitude 0.005 to the position $\bm p$ and rotation matrix $\bm R$, 0.035 to the velocity $\bm v$ and angular velocity $\bm \omega$, and 0.018 to the tilt angle $\varphi$.
\end{document}